# Analysis of Individual Conversational Volatility in Tandem Telecollaboration for Second Language Learning

Alan F. Smeaton[1], Aparajita Dey-Plissonneau[2], Hyowon Lee[1], Mingming Liu[1], Michael Scriney[1]

[1]Insight SFI Research Centre for Data Analytics , Dublin City University, Glasnevin, Dublin 9, Ireland.

[2]School of Applied Languages and Intercultural Studies, Dublin City University, Glasnevin, Dublin 9, Ireland.

alan.smeaton@DCU.ie

**Abstract:** Second language (L2) learning can be enabled by tandem collaboration where students are grouped in video conference calls while learning the native language of other student(s) on the calls. This places students in an online environment where the more outgoing can actively contribute and engage in dialogue while those more shy and unsure of their second language language skills can sit back and coast through the calls. We have built and deployed the L2L system which records timings of conversational utterances from all participants in a call. We generate visualisations including participation rates and timelines for each student in each call and present these on a dashboard. Students can self-reflect and perhaps target improving their levels of engagement for subsequent calls. We have recently developed a measure called personal conversational volatility for how dynamic has been each student's contribution to the dialogue in each call. This measures whether a student's contribution was interactive with a mixture of interjections perhaps interrupting and agreeing with others combined with longer contributions, or whether it consisted of regular duration contributions with not much mixing. We present an analysis of conversational volatility measures of a sample of 19 individual English-speaking students from our University at lower intermediate-intermediate level (B1/B2) in their target language which was French, in each of 86 tandem telecollaboration calls over one teaching semester. Our analysis shows that students varied considerably in how their individual levels of engagement changed as their telecollaboration meetings progressed. Some students got more involved in the dialogue from one meeting to the next while others did not change their interaction levels at all. The reasons for this are not clear from the data we have and point to a need for further investigation into the nature of online tandem telecollaboration meetings. In particular there is a need to look into the nature of the interactions and see if the choices of discussion topics were too difficult for some lower intermediate students and that may have influenced their engagement in some way.

**Keywords:** Language learning, telecollaboration, conversation metrics, conversational volatility.

## 1. Introduction

Second language learning (L2) can be supported by a pedagogical approach known as tandem telecollaboration where students are grouped into video conference calls while learning the native language of other student(s) on the calls (O'Dowd, 2018), (Wang and Wang, 2019). The first half of such online calls are carried out in the native language of the first group of students which is also the target language of the second group, and for the second half of the call the language used in the conversation is reversed and becomes the native language of the second group and the target of the first. Tandem telecollaboration has become popular over the last decade and especially very recently in our university as a result of teaching and learning having pivoted to more online activity as a result of the COVID pandemic.

Tandem telecollaboration places students in an online environment where more outgoing students can actively contribute and engage in the conversation while those more shy and unsure of their L2 language skills can sit back and coast through the calls but only if they are also partnered with a local fellow student. If a shy student is the only one from a given institution then s/he is forced to talk and to interact more. We have built and deployed the L2 Learning system which records the exact timing of all conversational utterances from all participants. This timestamped transcription of calls can be used to generate a range of metrics on participants' levels of interaction and contribution and this itself becomes a useful tool for students to reflect on their language learning during those online calls. The L2 Learning system has been in use in our university and in partnerships with seven other Universities, supporting hundreds of students in our University learning French, Spanish, German, or Italian as second languages, and students in Universities in those countries learning English as a second language (Dey-Plissonneau et al, 2021a). In total almost 1,000 hours of recorded online meetings have been analysed, visualised and used by students to support their L2 learning.

In the L2 Learning system we generate visualisations as shown in Figure 1 below including participation rates and interactive timelines for each student's participation in each online call and we present these on an online dashboard, personalised for each student. Students can self-reflect on their language learning by playing back various parts of the video recordings of their calls and navigating to parts where their contributions were the greatest, or the parts where they were silent. They can then use this self-reflection to target improving their engagement levels for subsequent calls.

As part of the analysis of student engagement in their tandem telecollaboration calls, we include a measure called conversational volatility which is a quantification of how dynamic and interactive a recorded call has been. This is visualised on the online dashboard and allows students to see an indication of the level of interaction among participants in a meeting before commencing playback of that online meeting. Calls with high levels of conversational volatility tend to be highly interactive and animated, with multiple student contributions with short duration affirmative confirmations and agreements mixed with longer contributions with lots of interruptions from others. Such animated conversations whether held online via video conferencing or face to face, are engaging and usually enjoyable for participants, and we hypothesise will probably lead to better language learning.

We have recently extended the measure of conversational volatility as applied to a whole meeting, to a measure called personal conversational volatility representing how dynamic has been each student's individual contribution to the dialogue in each online call. This measures whether a student's contribution was interactive with a mixture of interjections perhaps interrupting and agreeing with others combined with longer contributions, or whether it consisted of regular long, or short, contributions with not much mixing.

We present an analysis of conversational volatility measures of 19 individual native English speaking students from our University who are learning French, in each of more than 86 tandem telecollaboration calls over one teaching semester. This subset of students is an extract of just one class/course and allows us to do a deeper dive into what the values of the measure tell us. We analyse 67 hours of Zoom recordings and investigate how the levels of engagement in conversations by students who are at lower intermediate to intermediate levels in their target language, change throughout the semester as they take part in weekly telecollaboration calls. We can examine how many actually improve their interaction levels and by how much. Our analysis also shows how participation rates in those calls improve throughout the semester as students gain confidence in their own conversation abilities as they converse with native speakers of their target language.

The rest of this paper is organised as follows. In the next section we present a brief overview of the L2 Learning system, followed by an introduction to the nature of turn-taking in spoken dialogue and the conversational volatility metric previously introduced. We then present the analysis of our results followed by conclusions.

## 2. The L2 Learning System

The L2 Learning system is a web-based platform that allows students to upload the recording of their online tandem telecollaboration meetings, scheduled as part of their learning of a second language, and is described in Dey-Plissonneau et al (2021b). The process for students is as follows. Having registered on the L2 Learning system, students use the system to register a forthcoming meeting and are issued with a unique meeting code which is shared among all participants in that online meeting. They then use the Zoom platform to host and record their meeting. Importantly, the student hosting the Zoom meeting will have set her/his Zoom recordings to record to the cloud and to generate a recorded transcript of the meeting.

Some time after the meeting is completed, the hosting student receives email notifications from Zoom that their video recording and their audio transcript is available, the latter as a file in VTT format. VTT refers to Web Video Text Tracks (WebVTT) format and it contains supplementary metadata about a video file, in this case a timestamped transcript of the spoken audio, generated by Zoom. The hosting student then uploads links to the Zoom video recording and the VTT file to the L2 Learning system and within a few minutes all student participants in the meeting receive notification and a link to the visual dashboard for that meeting.

A screengrab of a sample meeting is shown in Figure 1 indicating a video playback window in the centre of the screen with 3 participants whose faces are blurred, and a timeline along the top of the screen with one colour (blue, red and yellow) for each participant. This shows when and where each participant contributed to the dialogue (clearly participant 1 in blue had more to say than the others),  Clicking any point on this timeline jumps the video playback window to that point in the video and commences video playback from that point, so the timeline is both for visualisation and video navigation.  The dashboard also includes a summary of each participant's amount of participation (shown as a percentage to the right of the timeline and also as a pie chart on the bottom centre), and a chord graph of conversation flow illustrating the ordering of speaker turn-taking.

The final component of the dashboard is conversational volatility which is described in the next section.  On the dashboard this appears as calculated for the entire meeting and for the first and second halves of the meeting as a bar chart (in blue) on the bottom left of the screen reflecting the French-speaking and English-speaking halves of the meeting respectively. This was introduced by Del-Plissonneau et al. (2022) and is a useful measure for students to support them as they reflect on an entire meeting. However it is not personalised in any way so in this paper we introduce conversational volatility for individual meeting participants, which we describe in the next section.

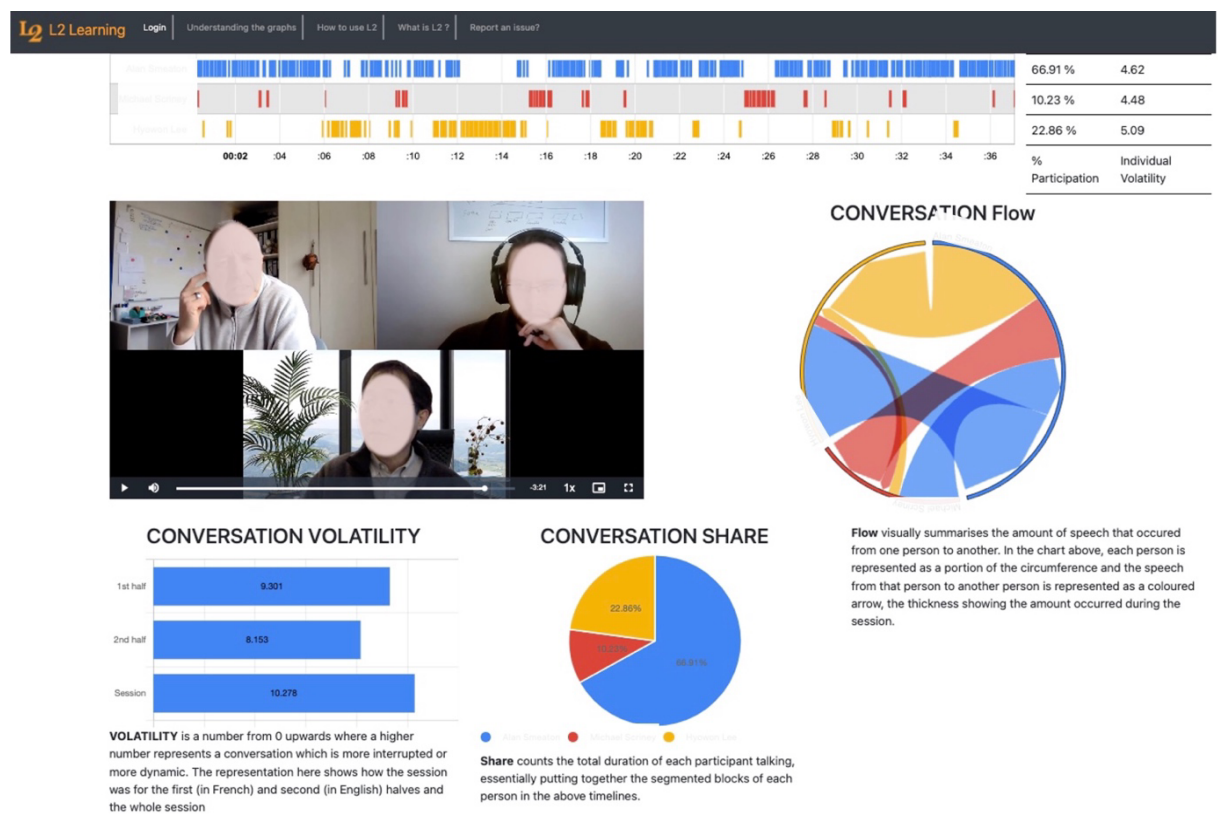

**Figure 1:** Screengrab of the visualisation dashboard for the L2 Learning system showing the analysis of an online Zoom call.

## 3. Conversational Volatility and the L2 Learning System

Research into the nature of multi-participant dialogue has been ongoing for decades but like many other aspects of computational linguistics has been accelerated by the availability of corpus data for researchers to work with. The largest example of this, published by Lowe et al (2015) contains more than 7 million utterances from multi-participant dialogues representing more than 100 mullion word occurrences from text chats rather than from aural conversations. Conversations are our focus and the differences between written and spoken language has been well explored, for example by Wallace and Tannen (1987). Researchers have tried to model the nature of spoken language and dialogue, ultimately with the aims of understanding it and being able to build computational systems that mimic it accurately.

The role of short utterances in online video conversations is especially interesting to us here as we seek to explore how students engage on online Zoom conversations. Li et al (2014) report a quantitative study of three specific short utterances, namely *that's right*, *that's true*, and *that's correct*, and their variants. Their findings are that use of such phrases are to accept, to assess or appreciate a statement made by another participant, or to give an affirmative answer to a point made or a question asked by another participant. These are all examples of engagement and involvement in the dialogue. Earlier work by Edlund et al (2010) studied very short utterances and what constitutes them and prior to that, Traum and Heeman (1996) had sought to define a consensus on what defines a short utterance, concluding that it is from a single speaker without interruption, constitutes a single turn, is complete, defines a single speech act and is an intonational phrase, with a pause separating it from the next utterance.

In the work we report here, the definition of an utterance is that it is a contribution to an online video conversation which is of long enough duration and distinct enough to have been picked up by an automatic speech transcription facility and the utterance attributed to a change of speaker. Thus this conforms to Traum and Heeman's 1996 consensus definition of an utterance. Whatever definition we use however, assumes that by contributing an utterance to a conversation, a participant is demonstrating their continued engagement in that conversation, even while not actively speaking and that is what we want to measure.

While we focus here on analysis of the audio, or more correctly an analysis of the audio transcript, there are other non-verbal cues which participants use in everyday dialogue. In face to face conversations we use gestures and eye contact although these do not transfer easily to the videoconference environment. In Ishi, Ishiguro and Havita (2017) the authors explore the relationship between head motion such as nodding or shaking, and speech in multi-speaker free dialogue conversations and these non-verbal cues are present in online videoconference meetings when participants' cameras are switched on. They found that head motions may convey information as a form of backchannel and that the strength of this and its usage depends on the prior relationships among speakers, including familiarity with each other. It would be interesting to extend our work to capture non-verbal clues from analysis of the video stream and use this to measure engagement level following similar work by Lee et al (2021) which has been reported to measure engagement levels of students in Zoom lectures but that is outside the scope of this paper.

Historical volatility (Ederington and Guan, 2006), is a well-known measure in statistics with common applications in economics and finance, especially in predicting time series values such as those associated with stock markets where it is used in assessing financial investment strategies. . It is formally defined as the degree of variation of values of some continuous time series over time, usually measured by the standard deviation of changes in values such as daily stock prices.

In this paper we apply the historical volatility measure to turn-taking in dialogue from telecollaboration meetings and we call it conversational volatility. The rationale behind using it is that values of conversational volatility will indicate whether the dialogue was truly interactive and composed of mixtures of shorter and longer utterances, such as when people interrupt each other, or whether it consisted of long monologues with likely tedious turn-taking.

Conversational volatility for the dialogue from entire meetings was introduced in Del-Plissonneau et al. (2022) and applied to the dialogue contributions from all participants in a tandem telecollaboration call, and we refer to this as meeting conversational volatility (m-CV). Individual speaker conversational volatility (i-CV) is defined in a similar way to m-CV. It is the standard deviation of the change in duration between adjacent utterances except that for i-CV we calculate this for the utterances from each speaker in turn whereas for the meeting it was calculated for the utterances from all speakers.  The relationship between the value for m-CV and the i-CVs values for the individual speakers during that meeting is unpredictable and depends entirely on the composition of the meeting and the participation of the contributors and their levels of interaction and engagement in the meeting. Where any participant contributes a the meeting dialogue with short duration utterances or affirmations such as "Yeah", or "Umm" or "Oui" or disagreements such as "Naw" or "Arrgh" and that participant mixes those with longer duration and thus more meaningful contributions, then that individual's i-CV will be high. Correspondingly, for a participant who engages only with short duration affirmations, agreements or disagreements, or who engages only with long duration contributions and then stays out of the conversation, then that individual's i-CV will be low.

To demonstrate how CV values for participants in the same meeting can vary we will illustrate by example, in the style of Anscombe's quartet (Anscombe, 1973).  Anscombe's quartet is a famous example in data science of a collection of 4 datasets with almost identical descriptive statistics but when visualised they reveal very different distributions of their values. In Figure 2 we show a fragment of a Zoom meeting between 2 participants, labelled Speaker 1 and Speaker 2. The top row of grey blocks indicates both the times and durations when Speaker 1 spoke and the second row indicates the same for Speaker 2. The row of numbers at the bottom of the Figure shows the durations of each utterance from each speaker in seconds, so it shows 4 seconds for Speaker 2 followed by 2 seconds for Speaker 1 followed by 2 seconds for Speaker 2, etc.  The values shown are for a 60 second clip taken from a recording of an actual interaction on a recorded 2-person Zoom meeting, with the values of the utterance durations rounded to the nearest second.

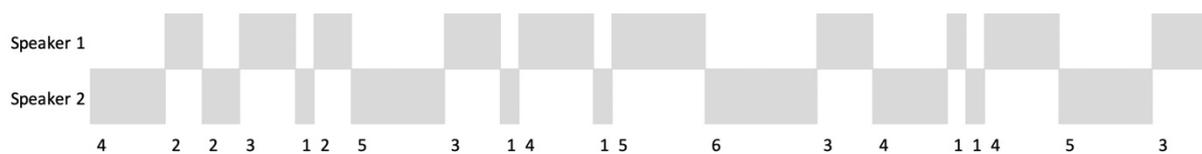

**Figure 2:** Illustration of turn taking and durations of contributions from 2 speakers in an extract from a sample Zoom meeting among two colleagues.

From a visual inspection of Figure 2 we can see that each of the speakers spoke for the same overall duration, 30 seconds each, and each speaker's contributions varied from just 1 second (probably an affirmation or a disagreement with a point being made) and a maximum of either 5 or 6 seconds.  Speakers also made the same number of contributions (10 each, 20 in total) to the 60-second clip so at first glance the two speakers would appear to be equal in terms of their contributions to the discussion. This is summarised in Table 1. However, exploring this more deeply through individual CV values reveals subtleties not immediately apparent.

**Table 1:** Conversational volatility measures for an extract from a sample Zoom meeting

|  | Speaker 1 | Speaker 2 |
|---|---|---|
| Duration of overall speech | 30 seconds | 30 seconds |
| Individual Conversation Volatility (i-CV) | 1.09 | 2.44 |
| Meeting Conversational Volatility (m-CV) | 1.42 ||

In Table 1 we see the computed conversational volatility measures for each individual speaker (i-CV) and for the whole of the sample 60-second clip including all speakers from the student meeting (m-CV).  This indicates that the value of Speaker 2's i-CV is more than twice that of Speaker 1, and that m-CV lies between the two i-CV values, but not at the mid-point. Higher values of conversational volatility, either individual or on entire meetings, correspond to a greater mixture of longer and shorter duration contributions. This short example illustrates the value of adding individual conversational volatility as feedback to student participants in tandem telecollaboration, so they can reflect on what the meeting was like as a whole (m-CV) and also on their own contributions and the contributions of others to the meetings (i-CVs).

## 4. Results and Analysis

The 19 students from our University, chosen for this analysis are native speakers of English and are learning French at lower intermediate to intermediate levels. They conduct telecollaboration meetings with students from a University in France who are native speakers of French and who are learning English. Table 2 presents the characteristics of the data used.

**Table 2:** Characteristics of dataset used

| Number of students | 19 |
|---|---|
| Number of Zoom meetings | 86 |
| Total duration of meetings | 66 hours, 57 minutes |
| Average duration of meetings | 47 minutes |
| Average number of participants per meeting | 3 or 4 |
| Average speaking time per student for entire meeting | 38% |

All of the 86 Zoom telecollaboration calls with 1 exception had either 3 or 4 participants. The average speaking time for our 19 students in these calls was 38% and with an average of 3.4 participants this means that these students contributed more to the conversations in terms of their speaking time, than other participants, their peers from France. There is no real explanation for this except to note that the English-speaking students learning French tended to speak more than the French-speaking students learning English.

We computed conversational volatility for each meeting (m-CV) and our initial interest is in how this progresses from the first to last meeting. The average m-CV across all 86 meetings was 7.596 and the average m-CV for the first, second, etc. meeting across all students is shown in Table 3 while the distribution of those m-CV values is shown in Figure 3 as a box and whisper plot.

**Table 3:** Average conversational volatility for student meetings

|  | Meeting 1 (19) | Meeting 2 (19) | Meeting 3 (19) | Meeting 4 (16) |
|---|---|---|---|---|
| Average m-CV | 7.113 | 8.478 | 7.905 | 5.531 |

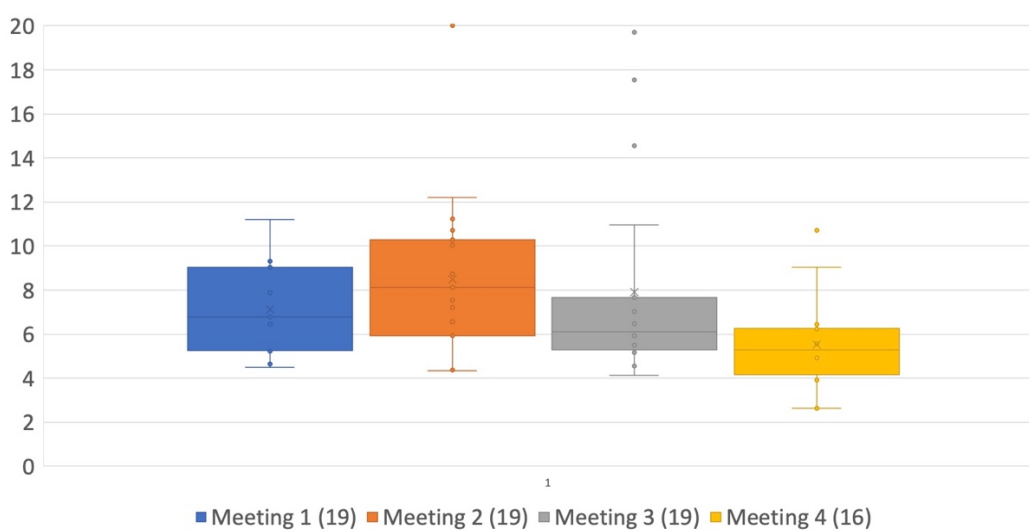

**Figure 3:** Distribution of m-CV values for first, second, third and fourth meetings for all students, y-axis shows conversational volatility value. Numbers in parentheses show the number of such meetings, 19 students had a first meeting, 16 students had a fourth meeting.

What we see from Table 3 and from Figure 3 is that the average meeting CV values across the meetings decreased, which is surprising. We expected the online meetings to become more interactive as students got to know each other given that the first time they would have met would have been at their first meetings, and this did not happen. This makes us wonder about the nature of those meetings so we calculated individual conversational volatility (i-CV) values for each of the 86 meetings our 19 students took part in. The average i-CV for the meetings was 3.967 so the absolute values of i-CV compared to m-CV are much less but this does not matter since when computing m-CV across a group of participants in a meeting this is more likely to see short utterances and interruptions which can come from any participant and thus increase the level of volatility measure compared to contributions from a single individual.

We are interested in how the i-CV values change as meetings progressed from first to last and whether those i-CVs values increased, decreased or stayed the same. To examine this, for each student we calculated the slope of the linear regression line through their i-CV values for all that student's meetings. This is formally defined as

$$b = \frac{\sum (x - \bar{x})(y - \bar{y})}{\sum (x - \bar{x})^2}$$

and its interpretation is illustrated in Figure 4 where we see two sample sets of data with slopes of +1.0 and -1.0 respectively.

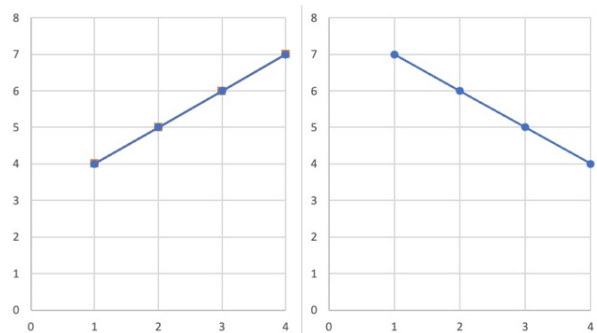

**Figure 4:** Sample calculations of slope with x- and y-axes showing sample values. Chart on the right has a slope = +1.0, chart on the left has a slope = -1.0

The distribution of slope values for the individual conversational volatility measures (i-CV) for our student cohort is shown as a box and whisper plot in Figure 5. This shows that the average slopes for entire meetings were positive (0.430) though for the French-speaking half of the meetings was negative (-0.471) and for the English-speaking half was slightly positive (0.126). The summary interpretation of this is that individual student engagement in their online tandem telecollaboration meetings did get more interactive as meetings progressed on average for meetings as a whole though less so for the French-speaking parts of their online meetings. However there were large differences in how some students engaged more and others less as their meetings progressed, as shown in Figure 5.

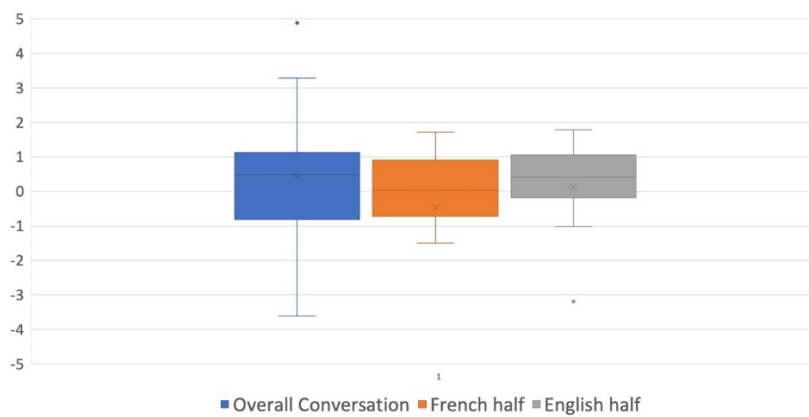

**Figure 5:** Distributions of slope values shown on y-axis for individual conversational volatility for entire meetings, for the first half of meetings (held in French) and for the second half of meetings (held in English).

## 5. Conclusions

In this paper we have presented an analysis of the results of automatically computing measures of students' levels of conversational interaction with fellow students in tandem collaboration as part of learning a second language. These measures were shared with students on an online dashboard during the semester though we are unsure what was the effect of having this information. Our analysis shows that students varied considerably in how their individual levels of engagement changed as their telecollaboration meetings progressed. Some students got more and more involved in the dialogue from one meeting to the next while others did not change their interaction levels at all, or became less engaged.

The reasons for this are not clear from the data we have and point to a need for further investigation into the nature of the meetings. Given that tandem telecollaboration is part of student-led autonomous learning we are unwilling to introduce closer monitoring of the meetings by lecturers so we take the approach of helping students to monitor their own interactions as part of their learning via the L2 Learning system and its dashboard.

However we are conscious that the choice of topics for discussion during the meetings where those topics were assigned to students, may have been either too difficult for some lower intermediate students or perhaps just not interesting enough for them. This may have influenced their engagement in some way, forcing them to be less engaged if they were not well versed or familiar with the discussion topic. For future usage of the L2 Learning system we will make more careful choices of discussion topics and perhaps make them student-led.

**Acknowledgement:**

The work reported in this paper was part-funded by Science Foundation Ireland (SFI) under Grant Number SFI/12/RC/2289_P2 (Insight SFI Research Centre for Data Analytics), co-funded by the European Regional Development Fund.

Student participants have given informed consent for us to use their data, as approved by our institution's research ethics board with reference DCU REC2021_205.